\documentclass[letterpaper, 10 pt, conference]{ieeeconf} 

\IEEEoverridecommandlockouts 

\usepackage{amsmath} 
\usepackage{graphicx}
\usepackage{amsfonts}
\usepackage{multirow}
\usepackage{multicol}
\usepackage{bbm}
\usepackage{cite}
\usepackage{color}
\usepackage{booktabs}
\newcommand\scalemath[2]{\scalebox{#1}{\mbox{\ensuremath{\displaystyle #2}}}}

\usepackage{dsfont}

\title{\LARGE \bf Visual-Kinematics Graph Learning for Procedure-agnostic \\ Instrument Tip Segmentation in Robotic Surgeries}

\author{Jiaqi Liu$^{1}$, Yonghao Long$^{1}$, Kai Chen$^{1}$, Cheuk Hei Leung$^{2}$, Zerui Wang$^{2}$, Qi Dou$^{1}$
\thanks{$^{1}$J. Liu, Y. Long, K. Chen, and Q. Dou are with the Department of Computer Science and Engineering, The Chinese University of Hong Kong. }
\thanks{$^{2}$C. H. Leung and Z. Wang are with Cornerstone Robotics Ltd.}
\thanks{This work is supported by Shenzhen Portion of Shenzhen-Hong Kong Science and Technology Innovation Cooperation Zone under HZQB-KCZYB-20200089, and Cornerstone Robotics Ltd.{\tt\small }}
\thanks{Corresponding author: Qi Dou (qidou@cuhk.edu.hk).}}

\begin{document}

\maketitle
\thispagestyle{empty}
\pagestyle{empty}

\begin{abstract}
Accurate segmentation of surgical instrument tip is an important task for enabling downstream applications in robotic surgery, such as surgical skill assessment, tool-tissue interaction and deformation modeling, as well as surgical autonomy. However, this task is very challenging due to the small sizes of surgical instrument tips,
and significant variance of surgical scenes across different procedures. Although much effort has been made on visual-based methods, existing segmentation models still suffer from low robustness thus not usable in practice. Fortunately, kinematics data from the robotic system can provide reliable prior for instrument location, which is consistent regardless of different surgery types. To make use of such multi-modal information, we propose a novel visual-kinematics graph learning framework to accurately segment the instrument tip given various surgical procedures. Specifically, a graph learning framework is proposed to encode relational features of instrument parts from both image and kinematics. Next, a cross-modal contrastive loss is designed to incorporate robust geometric prior from kinematics to image for tip segmentation. 
We have conducted experiments on a private paired visual-kinematics dataset including multiple procedures, i.e., prostatectomy, total mesorectal excision, fundoplication and distal gastrectomy on cadaver, and distal gastrectomy on porcine. 
The leave-one-procedure-out cross validation demonstrated that our proposed multi-modal segmentation method significantly outperformed current image-based state-of-the-art approaches, exceeding averagely 11.2\% on Dice.
\end{abstract}

\section{INTRODUCTION}

\begin{figure}[!t] 
\centering
\includegraphics[width=1\linewidth]{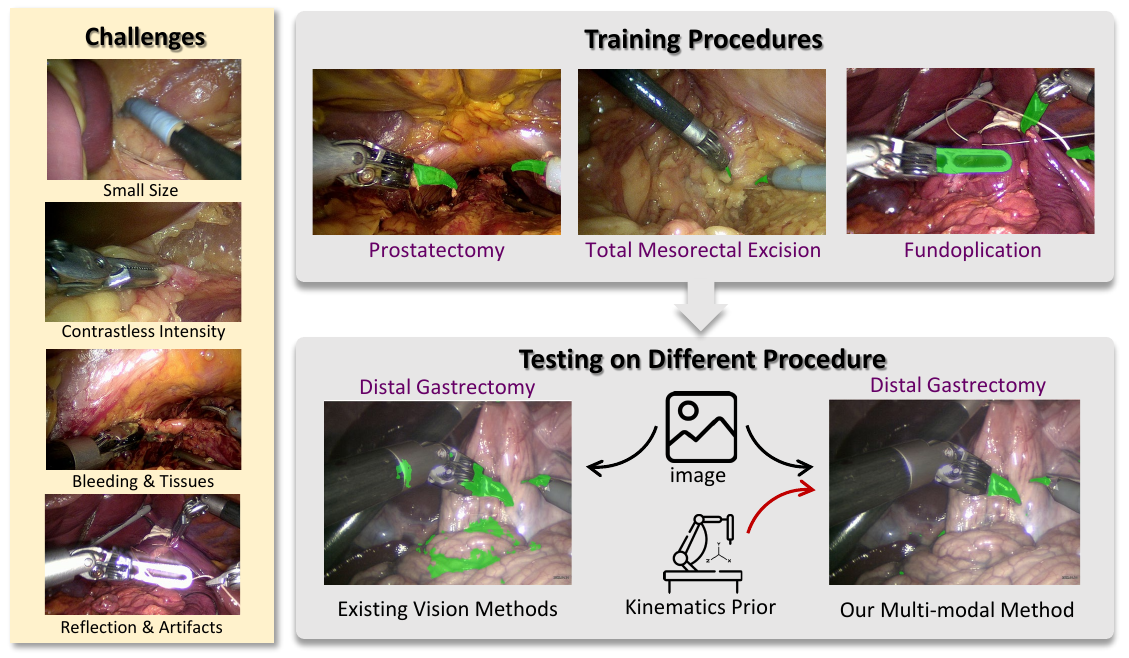}
\caption{The illustration of procedure-agnostic instrument tip segmentation. The left block lists the challenges for tip segmentation. The right block shows the procedure-agnostic settings, and comparison between our proposed method and existing vision based methods.}\label{fig::segmentation_examples}
\vspace{-4mm}
\end{figure}

Surgical instrument segmentation is a core task for surgical robot perception, which is fundamental for many downstream applications including futuristic ones such as skill assessment and robot autonomy. 
Despite extensive investigations on this task~\cite{allan20192017,allan20202018,shvets2018automatic,pakhomov2019deep,colleoni2020synthetic}, existing instrument segmentation solutions are still not precise enough for the tip part of the tools, nor robust across different types of surgical procedures (see Fig.~\ref{fig::segmentation_examples}).
This is currently the bottleneck for performance of state-of-the-art segmentation models, and is very difficult to be solved. Main challenges lie in the small size, contrastless intensity, various appearance due to different design and pose of the tool tip, as well as visual artifacts such as reflection and blood occlusion.
Furthermore, in surgery, the tool tip has frequent interactions with soft tissues, therefore, compared with segmenting other tool components such as the base and wrist, the instrument's tip segmentation is more prone to large variations of surgical scenes introduced by different types of procedures.




Previous works mostly studied pure image-based segmentation neural networks. Representatively, Iglovikov \textit{et al.}~\cite{iglovikov2018ternausnet} and Islam \textit{et al.}~\cite{islam2020ap} proposed to aggregate multi-scale context information to segment instrument into different parts. Jin \textit{et al.}~\cite{jin2022exploring} proposed to model intra-frame and inter-frame relation for improving the performance of part segmentation. Though promising results have been obtained, these methods cannot extract visual features with robustness, thus would be easily affected by the change of background scenes when applied to different procedures. To mitigate this problem, methods have been proposed to handle cross-procedure instrument segmentation. For instance, Sahu \textit{et al.}~\cite{sahu2020endo} proposed a consistency learning method for unsupervised domain adaptation of instrument segmentation. Zhao \textit{et al.} proposed a meta-learning based framework\cite{zhao2021one} to adapt the segmentation model at test time. Some works\cite{jin2019incorporating, sestini2023fun} aimed at adopting procedure-agnostic visual features such as optical flow and finally enhanced single-procedure performance. However, these methods cannot satisfy the requirement of real-time inference for robotic tasks, thus not usable in practice. In addition, and essentially, they still suffer from limited generalizability as pure vision models, and can hardly be procedure-agnostic for various surgeries.

Such limitations can be potentially addressed by incorporating kinematics information in the network. Different from vision, kinematics is scene-irrelevant as it only describes the status and properties of the robot (such as position, rotation, shape, etc.) which are more reliable and consistent across varying scenes. Accordingly, kinematics is more robust in preserving vital topological and geometrical priors of the surgical instrument and won't be affected by the environment. Very recent works~\cite{ding2022carts,ding2022rethinking,lu2023image} shared similar idea by using differentiable rendering offline to iteratively update the pose given the color image and kinematics rendered image for robust whole instrument segmentation. Beyond instrument segmentation, joint use of visual and kinematics embeddings has been explored for surgical action recognition~\cite{van2022gesture,long2021relational}. They propose a multi-modal relational graph network to learn different relationships between image and kinematics features, which demonstrates outstanding performance in terms of accuracy. Inspired by these successes, we consider borrowing the idea of multi-modal relational graph learning and make use of it for instrument tip segmentation. 

Specifically, accurately segmenting the instrument tip requires precise input information due to its significant challenges.  However, aligning kinematics perfectly to the image in pixel-level is not always possible due to unavoidable sources of deviations~\cite{richter2021robotic}, making pixel-wise fusing of image and kinematics information impractical. Recently, contrastive learning\cite{you2020graph, xu2021infogcl} has emerged as a powerful technique for cross-modal representation learning, which shows remarkable performance in describing high-level features from different modalities. In this regard, we are motivated to exploit the multi-modal information from the visual-kinematics data using cross-modal contrastive graph for robust procedure-agnostic instrument tip segmentation. 


In this paper, we first propose a graph learning module to learn the topological priors between tip and other parts of the instrument in high-level representational space, with features extracted from a CNN-transformer encoder. Then, a mesh rendering module is used to generate rendered mask of instrument parts from the kinematics data, with a similar CNN-transformer encoder adopted to extract robust instrument  prior. Last, a node-wise contrastive loss is designed to incorporate the robust and common topological and geometric representations from the kinematics graph to visual graph to form our final procedure-agnostic algorithm. Our main contributions are summarized as follows:
\begin{itemize}
\item We propose a novel visual-kinematics graph learning framework for instrument tip segmentation. The proposed method is efficient and robust to perturbation from image and diverse surgical procedures. 
\item We propose a cross-modal node-wise contrastive loss to incorporate the robust topological and geometric priors from kinematics to the image features, which can further improve the accuracy of tip segmentation.
\item We establish a comprehensive instrument segmentation dataset containing 5 different surgical procedures with paired visual and kinematics data. Leave-one-procedure-out cross validation is conducted on our dataset, and experimental results show significant improvement over the vision-based segmentation methods.
\end{itemize}

 \begin{figure*}[!t]
 \centering
\includegraphics[width=1\linewidth]{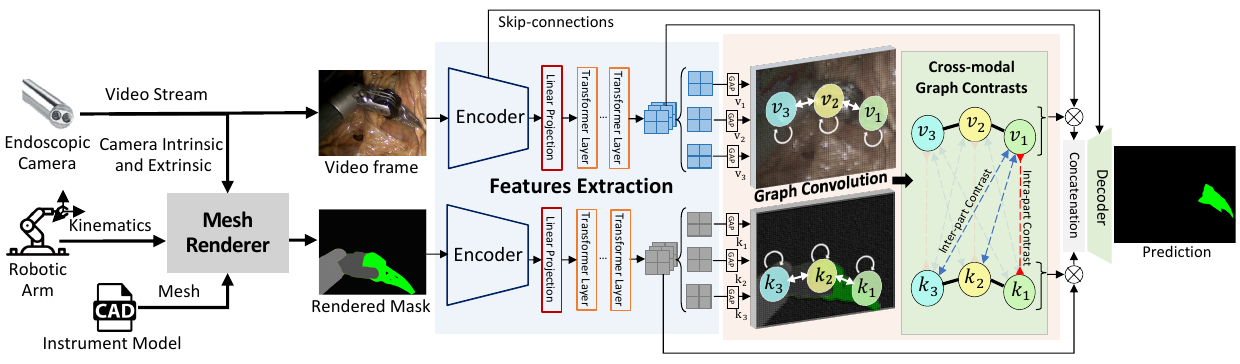}
\caption{The overall pipeline of our proposed visual-kinematics graph learning for procedure-agnostic instrument tip segmentation in robotic surgery.}
\label{fig::workflow}
\vspace{-2mm}
\end{figure*}




\section{METHODOLOGY}\label{sec::method}

The proposed visual-kinematics graph learning method for instrument tip segmentation is shown in Fig.~\ref{fig::workflow}. Our proposed method mainly consists of three distinct components. In section~\ref{sec21}, we describe the graph learning framework to model the instrument part relations in images. In section~\ref{sec22}, we introduce rendering of the robust instrument part-prior mask by kinematics data. In section~\ref{sec23}, we propose a node-wise contrastive loss to align the visual-kinematics part relations. And we finalize our optimization strategies in section~\ref{sec24}. Overall, this approach leverages instrument part priors and visual-kinematics contrasts for robust and accurate tip segmentation for procedure-agnostic surgeries.



\subsection{Instrument Part Relation Modeling with Graph Learning}\label{sec21}

We first describe the graph constructing for part relation modeling in images. Compared with directly segmenting the tip region in the image, the other parts in the instruments can provide essential topological information for a better understanding for the instrument tips. Specifically, 
the part relations for the instrument tips from the other instrument parts contain the articulated joints topology, the geometric information and pose constraints. Naturally, there thus exist geometric relations between the instrument parts. And the semantic information of each part could be inferred from the following joint angle and pose relations of other parts. 

In this regard, we develop a graph to model the part relations for the visible instrument parts in the image. Specifically, we model the relations of three instrument parts: the tip, the wrist and the base. Different from the previous method which built a causal graph\cite{ding2022carts}, we build a directed graph for propagating features along the geometric connections from the instrument base to tip. Such a model can explicitly model the relationship between the instrument joints in an efficient way. This design is also very suitable in our case, because geometric relationship is the core thing that needs to be modelled, which may not necessarily require a very complex model.


We incorporate such graph relations along the following steps. First, the input image $I \in \mathds{R}^{3 \times H\times W}$ will be forwarded to an encoder to extract the feature embedding $F_I \in \mathds{R}^{C \times h\times w }$.  Each encoder sequentially consists of a ResNet\cite{he2016deep} encoder and several transformer layers~\cite{dosovitskiy2020image} to extract the local and non-local features respectively. Second, the graph $G$ is constructed to model the geometric connections of the instrument parts. $G=\{V, E\}$,  $v_i \in  V$ containing the tip, wrist and base of the instrument, and $e_i \in E$ is connected as the instrument topology. Three layers of GCN~\cite{kipf2016semi} are adopted and each layer of the graph is updated as: 

\begin{equation}
    h_i^{l+1} = \sigma(\sum_{j \in E_i}(\frac{1}{c_{ij}} h_j^l W^l)),
\end{equation}

{
\setlength{\parindent}{0cm}
    where the $h_i^l \in \mathds{R}^{C/{|V|}}$ is the hidden state of the node $v_i$ in the $l$-th graph convolution layer. $E_i$ is the node indices which connected to node $v_i$, $W^l$ is the $l$-th layer graph parameters, $c_{ij}$ is the normalization constant for the edge $\{v_i, v_j\}$, and $\sigma$ represent the graph convolution operation and activation function, respectively.  And in detail, the initial hidden state of each node is from the partitions along the channels of the high-level embedding $F_I$. Global average pooling (GAP) is used for these partitioned embeddings to reduce the feature complexity. 
    Meanwhile, deep supervisions are adopted for each partition to ensure the graph node embedding could match with the corresponding instrument part. Next, we stack the learned hidden features $h_i$ in all the nodes together along the partitioned order. Finally, the stacked hidden feature $h$ would take effect as the channel-wise attention to the original feature map $F_{I}$, and the fused feature map would be adopted at the following decoding stage. 
}





\subsection{Robust Part Priors Generation by Kinematics Rendering}\label{sec22}


The graph learning framework mainly modeled the instrument part geometric relations from the corresponding visual context. However, such a relation trained from limited surgical procedures does not cover sufficient visual context, and such limited context cannot handle the wide visual changes in unseen surgical scenarios, thus leading to performance drop in testing. Kinematics data from surgical robots are considered robust across surgical procedures since its recording has no relation with the conducted surgeries, but it presents challenge for ways to adopt kinematics data. 

Inspired by existing kinematics-based segmentation methods that use kinematics data\cite{ding2022carts, da2019self, sestini2021kinematic} to iterative reach the true position of the robotic instruments, we noticed that the main relations between the visual-kinematics data are the geometric relations. And such relation could be regarded as prior information for the instruments in image. In detail, though the kinematics data suffer from the unavoidable deviations, the joints angles and the pose coordinates recorded in kinematics can be considered consistent with those in images. As a result, we argue that the direct modeling of the high-level part relations between the visual-kinematics data instead of iterative optimization could also help to improve the segmentation quality for the unseen surgeries.

Compared with simulating all the scenes during surgeries or rendering all the clear objects to 2D images, rendering the silhouette mask of the parts in instruments provides enough part-priors for extracting the geometric relations. Thus, the kinematic rendering module is designed to generate silhouette masks with the robust part-prior information. We generate $R\in \mathds{R}^{H \times W}$ by image rendering. The kinematics data containing the $x$,$y$,$z$ and its quaternions $K_i \in \mathds{R}^{n_s\times 3 \times 7}$ and the instrument CAD model of total $n_s$ instruments are used to render the meshes $M$, where each part is painted with a unique color. Then we adopt the Open3D~\cite{zhou2018open3d} to render the mask $R \in \mathds{R}^{H\times W}$ by a pinhole camera model, where kinematics of the camera is $K_c  \in \mathds{R}^{1 \times 3 \times 7}$. To speedup the single rendering process, the mesh decimation strategy is applied to generate more lightweight meshes $M_d$ from $M$. And besides, a scaled rendered mask $R_s \in \mathds{R}^{\frac{H}{s}\times \frac{W}{s}}$ is adopted to reduce the rendering computation cost, then $R_s$ is resized to the original size. The two strategies finally led to a 4 times speed up with trivial accuracy decrease. The process is formulated as below:
     
\begin{equation}
 R = Resize(f_{RM}(M_d, K_i, K_c)),
\end{equation}

{
\setlength{\parindent}{0cm}
where the $f_{RM}$ represents the rendering module. 
}
The rendered mask would be put into a similar graph learning framework as~\ref{sec21}, and a graph $G'$ with the same structure will be finally constructed to provide part-priors for the paired visual data.

\begin{figure*}[!t]
\centering
\includegraphics[width=1\linewidth]{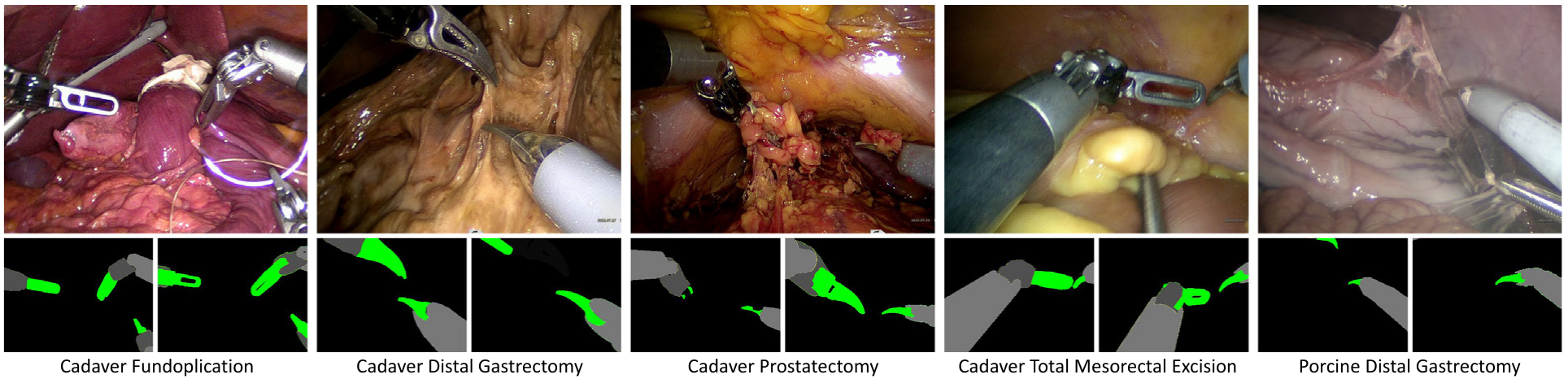}
\caption{Examples of 5 different surgical procedures from our proposed dataset. Second row shows labels (left) and the kinematics rendered masks (right).
}\label{fig::dataset}
\vspace{-1mm}
\end{figure*}

\subsection{Cross-modal Relation Modeling via Contrastive Graph}\label{sec23}
    

 Next, to incorporate the robust part-prior relations from kinematics $G'$ to the image graph $G$, we explore the graph contrastive relations for a better fusion. Sole graph structures with direct fusing cannot guarantee the propagated geometric relations are aligned between the visual-kinematics data. Furthermore, the learned features of graph nodes may be confused with each other because of limited inter-class variations, and such a confusion could ultimately interfere with learning clear geometric relations.

    
To tackle this issue, we propose a node-wise contrastive loss to model the visual-kinematics relations. Since our extracted relations for a single modality are mainly based on part relations and geometric connections, two relations are required for the two graphs: (1) The same node from the two graphs should share similar representation, indicating that the same part of the instrument between the two modalities should have similar relations. This relation would improve the modeling of a single part. (2) Different nodes from the two graphs should have different representations, reducing the ambiguity of the geometric connection of different parts. This relation would improve the discrimination of features and the modeling of different parts. Furthermore, since the two relations are satisfied, the relations described by the two directed graphs would tend to be similar. To describe the relations, we define the node-wise contrastive loss as follows:

{\small
\begin{equation}
\begin{aligned}
L_{nc} =  {\frac{1}{|\mathds{V}|}} \sum_{ h_i \in  \mathds{V}}
          -log {\frac{exp(h_i \cdot h_i^{+} / \tau)}{exp(h_i \cdot h_i^{+} / \tau) + \sum_{ h_j^{-} \in N} exp(h_i \cdot h_j^{-} / \tau)}}, \\
\end{aligned}
\end{equation}
}

{
\setlength{\parindent}{0cm}
    where, $h_i$ is the hidden feature of each node, $h_i^+$ represents the positive features from the same node of another graph, $\mathds{V}$  is all the nodes of the two graphs $G$ and $G'$, and $h_j^-$ represents the negative features from the different nodes $N$ of another graph. $\tau$ is the temperature.
}

Once the geometric relations between visual-kinematics has been defined, we finally aligned the geometric relations of the two modals. As observation above, since kinematics data is robust across different surgical procedures, the tip segmentation would ultimately benefit from the geometric information from the part-prior given by the kinematics data, resulting in a more robust performance for procedure-agnostic segmentation. Our experiments~\ref{quali_res} also confirm the usefulness of the defined relations.

\begin{figure*}[!t]
\centering
\includegraphics[width=1\linewidth]{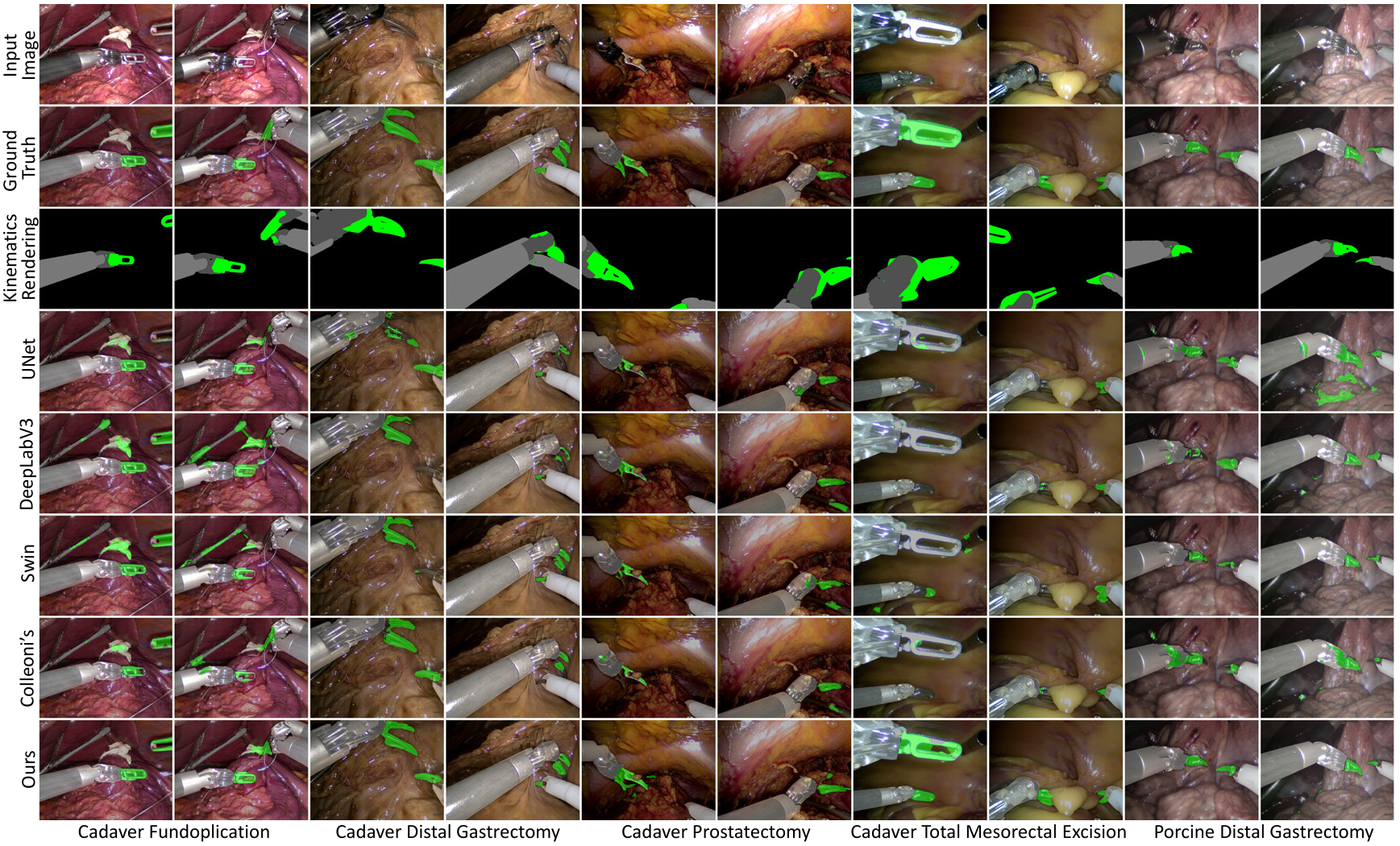}
\caption{Qualitative Results of the proposed method and baselines. Images from different procedures are selected. We mainly focus on the difficult cases such as non-instrument tool interference, tissue interaction, small size and reflections. }\label{fig::quali}
\vspace{-2mm}
\end{figure*}

\subsection{Overall Training Process}\label{sec24}
Overall, two tasks are considered. The main task is aimed to estimate a better low-level pixel-wise classification, while the auxiliary task is our proposed node-wise contrastive loss. Given the input data $I$, $R$, and the corresponding segmentation mask $y$, the model is trained to minimize two loss jointly by optimizing the following:
\begin{equation}
\scalemath{0.91}{
\begin{aligned}
L_{train} = & L_{seg}(I,R; y) + \lambda_1 L_{nc}(F_I, F_R) + \lambda_2 L_{ds}(F_I, y),
\end{aligned}}
\end{equation}

{
\setlength{\parindent}{0cm}
where $\lambda_1$ and $\lambda_2$ are hyper-parameters to balance the losses. The segmentation loss $L_{seg}$ is the CEDiceLoss for the three parts, while the deep supervision loss $L_{ds}$ is the binary CEDiceLoss for each single part. Test-time batchnorm updates strategy \cite{wang2020tent, ioffe2015batch} is adopted to reduce image variance.
}








\section{EXPERIMENTS}\label{sec::exp}
\subsection{Multi-Procedure Visual-Kinematics Dataset}

To verify the effectiveness of the proposed method, we collected a challenging dataset during surgical trials using a private surgical robot system. Our dataset consists of 5 sequences of different surgical procedures, with a total of 950 images. The procedures include porcine distal gastrectomy, cadaver distal gastrectomy, cadaver fundoplication, cadaver prostatectomy, and 
 cadaver total mesorectal excision. Ethics approval of the study protocol performed at Multi-Scale Medical Robotics Center (MRC) has been granted by IACUC of the Hong Kong Science and Technology Parks Corporation (HKSTP) with a reference number (HKSTP IACUC ref. no.: 2021-010). We consider the two procedures named distal gastrectomy to be different because the surgical phases and environment between porcine and cadaver are different. For each procedure, we selected sequences in which significant instrument interactions were observed from independent videos. To address the discriminative scenarios, each sequence contains different scenarios, such as tissue occlusion, extreme close focus, smoke, and non-tool instrument interference. Each sequence is around 8 minutes long, with a frame rate of 0.5 fps. Each frame in the sequence has a resolution of 960$\times$1280. The kinematics raw data is recorded at a rate of 5 Hz and contains 3-dimensional position information and 4-dimensional quaternion rotation parameters. The kinematics sequence with the nearest timestamp matches the image frames. Frames without instruments are manually removed and extended sequentially~\cite{allan20192017,allan20202018}. Both left and right eye images from the stereo camera are provided.

    The dataset was annotated by an engineering student and verified by technicians of the surgical robot from the company. We labelled only the left frames to reduce the workload. The annotations are performed with the "LabelMe"~\cite{torralba2010labelme} toolbox. Three parts of the instrument are labeled, including the base, wrist and tip. Different types of surgical tools are included in our dataset. The illustration of the dataset, containing the images of different surgeries, annotated parts masks, and the rendered masks is shown in Fig.~\ref{fig::dataset}. It could be noticed that the rendered masks have similar geometric connection compared with the annotated masks, but they are shifted from the corresponding positions.

    This dataset is designed towards robust surgical tool parts segmentation with cross-procedure settings. In specific, each sequence has its specialized procedure and background knowledge. Besides, the paired kinematics data provides rich information about tool movements and the geometrical information, which could be used for multi-modal experiments.


\subsection{Implementation Details and Experiment Settings}

Overall, the entire pipeline processes the image data and kinematics data sequentially. In the rendering module, we set the decimation rate of meshes as 10 and the scaled factor as 2. The following framework can be trained end-to-end. We use the first 3 stages of the ResNet-34 for local feature extraction and 6 transformer layers for non-local feature extraction. The number of channels of high-level features is set to 768; hence, the dimension of the hidden state of graph modules is 256. The graph module contains 3 layers of GCN, and the decoder has 5 stages to decode the tip mask.

The graph module is implemented with Pytorch Geometric~\cite{fey2019fast} in PyTorch with an NVIDIA RTX 3090 GPU. The input images are resized to 960 $\times$ 960 and recovered to the original size after the network. We used the Adam~\cite{kingma2014adam} optimizer with learning rate of 1e-3 and StepLR scheduler with decay rate of 0.8 for every 5 epoch. $\lambda_1$ is set as 1 and $\lambda_2$ is set as 0.05. The maximum epoch is 100. During test, an NVIDIA RTX 3090 GPU is used for inference.

Our experiment setting follows the leave-one-procedure-out cross validation, to verify the effectiveness of the proposed model towards the unseen surgeries. 5-fold validations are conducted in total. The following evaluation metrics are employed: i) Dice Score. ii) IoU, the intersection-of-Union. The two metrics are designed to measure the intersection regions between ground truths and the predicted masks. iii) Inference fps, to evaluate the speed of proposed method. 

\begin{table}[!t]

\centering
\caption{Comparison with State-of-the-art Methods on Surgical Instrument Segmentation using Average Dice and IoU from Leave-one-procedure-out experiments.}\label{tab::quant}
\scalebox{0.9}{
\begin{tabular}{c|c|c|c|c|c}
\toprule
\multirow{2}{*}{Methods} & \multicolumn{2}{c|}{Input Data} & \multirow{2}{*}{mean Dice (\%)} & \multirow{2}{*}{mean IoU (\%)}  & \multirow{2}{*}{Fps}\\
\cline{2-3}
 ~ & Vis & Kin & & & \\
 \hline
UNet~\cite{ronneberger2015u} & \checkmark && 44.7$\pm$25.6 & 33.6$\pm$22.0 &16.6 \\
DeepLabv3~\cite{chen2017rethinking}&\checkmark&&43.4$\pm$26.6&32.2$\pm$22.8&7.2\\
Swin~\cite{liu2021swin}&\checkmark&& 41.8$\pm$25.3 &30.6$\pm$21.3& 4.9\\
Colleoni's~\cite{colleoni2020synthetic} &\checkmark&\checkmark& 37.1$\pm$24.4 & 26.3$\pm$19.9 &17.8\\
Ours & \checkmark&\checkmark&55.9$\pm$21.9 &42.8$\pm$20.4 & 14.4\\
\bottomrule
\end{tabular}}
\vspace{-2mm}
\end{table}

\begin{table*}[htbp]
  \centering
  \caption{Detailed Instrument Tip Segmentation Results on Each Individual Procedure for Different Methods.}\label{tab::seq_res}
    \scalebox{0.92}{
    \begin{tabular}{c|c|c|c|c|c|c|c|c|c|c}
    \toprule
    \multirow{2}{*}{Method} & \multicolumn{5}{c|}{IoU(\%)} & \multicolumn{5}{c}{Dice(\%)} \\
\cline{2-11}    & Proc1 & Proc2 & Proc3 & Proc4 & Proc5 & Proc1 & Proc2 & Proc3 & Proc4 & Proc5 \\
    \hline
    UNet~\cite{ronneberger2015u}  & 34.2$\pm$18.0 & 57.5$\pm$25.7 & 23.4$\pm$24.5  & 35.2$\pm$25.9 & 18.7$\pm$15.9 & 48.0$\pm$21.8 & 68.9$\pm$25.5 & 32.1$\pm$30.0 & 45.6$\pm$28.9 & 28.7$\pm$21.8 \\
    DeepLabV3~\cite{chen2017rethinking}  & 35.4$\pm$13.9  &  44.1$\pm$27.4 &  22.2$\pm$25.5 & 35.0$\pm$26.9  & 24.2$\pm$20.2 & 50.5$\pm$16.8 & 55.6$\pm$29.7 & 30.1$\pm$30.6 & 45.9$\pm$30.5 & 34.8$\pm$25.2\\
    Swin~\cite{liu2021swin}& 33.6$\pm$15.5  &  47.9$\pm$24.8 &  22.7$\pm$23.7 & 19.8$\pm$18.9 & 29.0$\pm$23.8 & 
    48.1$\pm$18.9 & 60.5$\pm$26.0 & 31.5$\pm$28.3 & 29.5$\pm$24.0 & 39.7$\pm$29.0 \\
\midrule
    Colleoni's~\cite{colleoni2020synthetic} & 36.2$\pm$15.7  &  37.6$\pm$22.1 &  19.0$\pm$23.6 & 18.8$\pm$21.8  & 20.8$\pm$16.3 &  
    51.1$\pm$18.2 & 50.7$\pm$24.9 & 26.2$\pm$29.5 & 26.5$\pm$26.7 & 31.4$\pm$22.2\\
    Ours & \textbf{47.8}$\pm$\textbf{13.6}  & \textbf{61.3}$\pm$\textbf{22.8} &  \textbf{32.7}$\pm$\textbf{20.2}  &  \textbf{41.2}$\pm$\textbf{24.7}  &  \textbf{30.8}$\pm$\textbf{20.9}  & \textbf{63.4}$\pm$\textbf{14.1}  & \textbf{73.0}$\pm$\textbf{21.6}  & \textbf{45.9}$\pm$\textbf{23.1}  & \textbf{53.7}$\pm$\textbf{26.7}  & \textbf{43.3}$\pm$\textbf{24.2}  \\
    \bottomrule
    \end{tabular}}%
  \label{tab:addlabel}%
  \vspace{-2mm}
\end{table*}%

\subsection{Quantitative Experiments}

We compare our methods with existing segmentation methods UNet~\cite{ronneberger2015u}, DeepLabV3~\cite{chen2017rethinking}, Swin~\cite{liu2021swin}, and Colleoni's~\cite{colleoni2020synthetic}. For a fair comparison, we use the same instrument part annotations when training the baseline methods to provide equal parts information. The overall results are shown in Table~\ref{tab::seq_res}. Our method has an obvious improvement for the unseen surgeries compared with the existing vision methods with 12.2\% improvement on mean Dice score and 9.4\% on mean IoU. This result identifed the effectiveness of fusing the kinematics data. In addition, our method is obviously improved compared with the existing multimodal method~\cite{colleoni2020synthetic} with the improvement at 18.8\% on Dice score , since directly fusion only works when the target has large intersection with rendering, such as the whole instrument. The high-level information, such as geometric relation learning would be more robust and ideal for fusing the deviated kinematics data. In this regard, our method provides a possible way for using the kinematics data as high-level priors or constraints. A notable overfitting could be noticed from the pure vision methods, since Swin has a more complex structure but results in a worse performance when testing at the unseen surgeries. Also, large standard deviations are recorded for all the methods in Table~\ref{tab::seq_res}, which can be caused by the types of tips from different instruments, the scales of tips and the frequent changes of tip interactions during surgeries.  However, after incorporating the kinematics information, the standard deviation becomes less. And our detailed results for each procedure could be referred at Table~\ref{tab:addlabel}. It indicates that our method brings a robust quantitative improvement for the different procedures in the dataset when compared with the existing methods. And these observations show that our method is robust for the procedure-agnostic settings. Also, our method achieves a good balance of performance and efficiency. The network and kinematics rendering module can run at 14.4 fps and 38.2 fps respectively,
satisfying the real-time requirement of surgical robot during clinical practice.  






\subsection{Qualitative Results}\label{quali_res}

In this chapter, we introduce the qualitative results cross procedures compared with the existing methods, which is shown in Fig.~\ref{fig::quali}. In general, our method shows a good performance of the tips with stable segmentation when the part relations between the visual-kinematics images could be approximately aligned. As illustrated, even though the rendered masks for every part are shifted, it could improve the final performance when the geometric relations are consistent from the images to the rendering masks. The tip regions can be well preserved across procedures even under the interference of the unseen tissues and unseen non-instrument tools when the tips have a severe interaction with the regions. The kinematics data contain the geometric information of each instrument, and this information would  help to distinguish the instruments from the other objects. However, other existing pure-vision methods confused the background and the tips, especially for the regions which have intersections with the tissues and non-surgical tools. Though the direct fusing of kinematics has less false positive, the tip regions are often missing since influence of the kinematic deviations. According to our observation, the segmentation performance actually has little relation with the accuracy of absolute position of the kinematics rendered mask, but the full acceptance of the three parts attaches importance. When some parts are missing in the rendered mask or in the image, the strategy of graph contrastive learning leads to a failure. The performance of our method will finally decrease.  

\begin{table}[t]

\centering
\caption{Ablation study for proposed method. }\label{tab::ab}
\begin{tabular}{c|c|c|c|c|c}
\toprule
 \multicolumn{4}{c|}{Modules} & \multirow{2}{*}{mean Dice (\%)} & \multirow{2}{*}{mean IoU (\%)}\\
\cline{1-4}
Vis& Kin & GCN & CL & \\
 \hline
 \checkmark&& & & 44.4$\pm$22.0 &32.0$\pm$19.0 \\
\checkmark&&\checkmark& & 46.9$\pm$20.5 &34.1$\pm$17.7\\
\checkmark&\checkmark&& & 48.5$\pm$21.9 & 35.5$\pm$19.2 \\
\checkmark&\checkmark&\checkmark&  & 54.4$\pm$21.0 &40.8$\pm$19.0\\
\checkmark&\checkmark&\checkmark&\checkmark& 55.9$\pm$21.9 &42.8$\pm$20.4 \\
\bottomrule
\end{tabular}
\vspace{-2mm}
\end{table}

\subsection{Ablation Study}

In this chapter, we analyze the contributions of each component of our proposed method.
The results of ablation study is shown in Tab.~\ref{tab::ab}. We compare four ablation studies of our proposed methods: (1) preserve the CNN-transformer encoder on visual branch, (2) preserve the CNN-transformer encoder and graph learning (GCN) module on visual branch, (3) preserve the CNN-transformers on the two branches, (4) preserve the CNN-transformers and the GCN module on the two branches without the contrastive learning (CL)  module.

From the experiments, we could notice that using the two modal data could provide richer information compared with only using the single visual modality. Obvious improvement can be observed when the kinematics data is fused with 4.1\% on Dice without the GCN structure and 7.5\% with the GCN structure. GCN also successfully modeled the geometric part relations of the instrument, and contribute to a better performance with a improvement of 5.9\% on Dice for the visual-kinematics module. This improvement indicates that the GCN has the power to model the part relations and aggregate the joints information on the tool segmentation task. Finally, the contrastive learning module further improved the performance as aligning the geometric relations. And the contrastive loss gives a 1.5\% improvement on dice score and 2.0\%  on the IoU. So, this result identifies the effectiveness of such a contrastive learning strategy.


\section{CONCLUSIONS}\label{sec::conclusion}
In conclusion, we have proposed a graph learning framework that can address the procedure-agnostic instrument tip segmentation task with the incorporation of kinematics-guided priors. The proposed framework achieved a superior performance across various types of surgeries. We successfully modeled the geometric part relations by constructing a simple graph as the joints connections. Then, we proposed a cross-modal node-wise contrastive loss to incorporate the robust topological and geometric priors from kinematics to the visual data, and finally improved the robustness of the algorithm. The collected cross-procedure visual-kinematics dataset validated our method. While this work focused on instrument tip segmentation, there is room for further investigation of leveraging more kinematics information for visual tasks to achieve better fusion performance. Moreover, one interesting topic remains to be studied is the robot-agnostic setting since many new surgical robotic systems with similar design are developed and commercialized. We hope more tasks can benefit from joint exploit of visual and kinematics information for surgical robotics.




\section{ACKNOWLEDGEMENT}\label{sec::ack}
We would like to thank Prof. Philip Chiu, Prof. Shannon Chan, Prof. Hon-chi Yip, Prof. Simon Ng, Dr. Man-fung Ho, Prof. Chi-fai Ng, and Dr. Samuel Yee for performing the pre-clinical trials.






\bibliographystyle{IEEEtran}
\bibliography{references}

\end{document}